\documentclass{article}

 \usepackage[preprint,nonatbib]{neurips_2023}

\usepackage[utf8]{inputenc} 
\usepackage[T1]{fontenc} 
\usepackage{hyperref}  
\usepackage{url}   
\usepackage{booktabs}  
\usepackage{amsfonts}  
\usepackage{nicefrac}  
\usepackage{microtype}  
\usepackage{xcolor}   
\usepackage{graphicx}

\title{Decomposing spiking neural networks with Graphical Neural Activity Threads}

\author{
 Bradley H. Theilman\\
 Cognitive and Emerging Computing\\
 Sandia National Laboratories\\
 Albuquerque, NM\\
 \texttt{bhtheil@sandia.gov}\\
 \And 
 Felix Wang\\
 Cognitive and Emerging Computing\\
 Sandia National Laboratories\\
 Albuquerque, NM\\
 \texttt{felwang@sandia.gov}\\
 \And 
 Fredrick Rothganger\\
 Cognitive and Emerging Computing\\
 Sandia National Laboratories\\
 Albuquerque, NM\\
 \texttt{frothga@sandia.gov}\\
 \And 
 James B. Aimone\\
 Cognitive and Emerging Computing\\
 Sandia National Laboratories\\
 Albuquerque, NM\\
 \texttt{jbaimon@sandia.gov}\\
}

\begin{document}

\maketitle

\begin{abstract}
A satisfactory understanding of information processing in spiking neural networks requires appropriate computational abstractions of neural activity. Traditionally, the neural population state vector has been the most common abstraction applied to spiking neural networks, but this requires artificially partitioning time into bins that are not obviously relevant to the network itself. We introduce a distinct set of techniques for analyzing spiking neural networks that decomposes neural activity into multiple, disjoint, parallel threads of activity. We construct these threads by estimating the degree of causal relatedness between pairs of spikes, then use these estimates to construct a directed acyclic graph that traces how the network activity evolves through individual spikes. We find that this graph of spiking activity naturally decomposes into disjoint connected components that overlap in space and time, which we call Graphical Neural Activity Threads (GNATs). We provide an efficient algorithm for finding analogous threads that reoccur in large spiking datasets, revealing that seemingly distinct spike trains are composed of similar underlying threads of activity, a hallmark of compositionality. The picture of spiking neural networks provided by our GNAT analysis points to new abstractions for spiking neural computation that are naturally adapted to the spatiotemporally distributed dynamics of spiking neural networks.
\end{abstract}

\section{Introduction}

The robustness, flexibility, and efficiency of natural intelligence is thought to emerge from the unique computational characteristics of the brain's highly recurrent and dynamical spiking neural networks. Despite an explosion of neuroscientific data at all organizational levels, our understanding of spiking neural computation is tentative. Both neuroscientists and researchers in neural computing seek the right abstractions. A theoretical framework is required to interpret the myriad of patterns of spiking activity generated by spiking neural networks and how these patterns relate to neural computations. In particular, spiking neural networks lack a satisfactory theory of compositionality, namely, how complex computations are built from simpler parts. 

Defining a useful abstraction for spiking neural computation requires specifying the meaningful relations between spikes that support these computations. Abstractions emerge from equivalences, and specifying a meaningful relation also specifies an appropriate notion of equivalence -- for example, the abstraction of binary bits in conventional computer architectures partitions the states of the physical system into the classes ``1'' and ``0''.

Most computational abstractions of recurrent spiking neural networks depend on the concept of a population state vector. In these approaches, simultaneous population activity forms a distributed representation of computational variables as a vector in a high-dimensional space \cite{vyas_computation_2020}. The action of the spiking network corresponds to a dynamical system evolving the state vector to the desired result of the computation.
The population state vector interpretation requires assuming that every neuron agrees that simultaneity with respect to the time bins is the meaningful relation between spikes that defines neural computations. If conduction delays between neurons are exactly equal or negligible, this may be valid. However, biological evidence suggests that delays and other asynchronous attributes of spiking networks are critical for computation \cite{egger_local_2020}. Regardless, the validity of reducing neural activity to a sequence of state vectors is an empirical question \cite{brette_philosophy_2015}. 

Another aspect of spiking networks is temporal variability. Spiking neural networks rarely respond to identical input with temporally identical spike trains, even though they are biophysically capable of such precision \cite{mainen1995reliability}. Within the state vector abstraction, this means that the computation must be expressed probabilistically, because the state vectors rarely trace identical sequences through time. Mounting evidence suggests that some neural variability is due to flexibly warping invariant spike patterns in time \cite{williams_discovering_2020} or through the unknown mechanisms behind representational drift \cite{rule_causes_2019}. Attributing neural variability to stochastic noise may overlook invariant patterns in less observer-centric descriptions. In other words, neural activity that looks variable may be exquisitely precise with respect to a different point of view \cite{mainen1995reliability}. 

Overall, the assumptions imposed by current abstractions of spiking networks are in tension with the observed properties of biological networks. This motivates a search for alternative descriptions of spiking neural networks that are more naturally adapted to spiking dynamics and can serve as a foundation for computational abstractions. In this work, we introduce an alternative approach to decomposing spiking neural network activity that avoids many of these assumptions. Our analyses reveal neural activity intrinsically partitions itself into disjoint causal threads of activity, that we refer to as Graphical Neural Activity Threads (GNATs). We analyze the GNATs that emerge from a spiking neural network simulation receiving both random and patterned stimulation through externally-imposed spikes. We also provide a technique for identifying analogous GNATs that reoccur at different times in our simulations. The GNATs display many properties that suggest they are a useful computational abstraction for spiking neural networks, such as spatiotemporal parallelism (more than one GNAT can exist in the same space or time) and compositionality (GNATs are built from smaller threads that are flexibly reused and rearranged). 

\section{Related Work}
Our definition of GNATs is related to the concept of polychronization \cite{izhikevich_polychronization_2006}. Polychronization is the observation that the simultaneous arrival of presynaptic spikes after heterogeneous axonal conduction delays leads to the formation of polychronous groups, or groups of spikes that repeat with precise temporal relationships. However, GNATs subsume polychronous groups and are more general. Furthermore, our algorithm for extracting analogous threads is more efficient than the original brute-force algorithms for extracting polychronous groups. 

GNATs also relate to the concept of cell assemblies. Cell assemblies are "transiently active ensembles of neurons" \cite{buzsaki_neural_2010} thought to support numerous neural computations. Traditionally, cell assemblies are defined by near-simultaneously active cells. GNATs generalize cell assemblies to transiently active collections of neurons arranged flexibly in time, bound together by the causal relations that define spiking dynamics, instead of temporal proximity with respect to an external clock.

\section{Graphical Neural Activity Threads}
\label{gen_inst}

\begin{figure}[h!]
\includegraphics[width=1\textwidth]{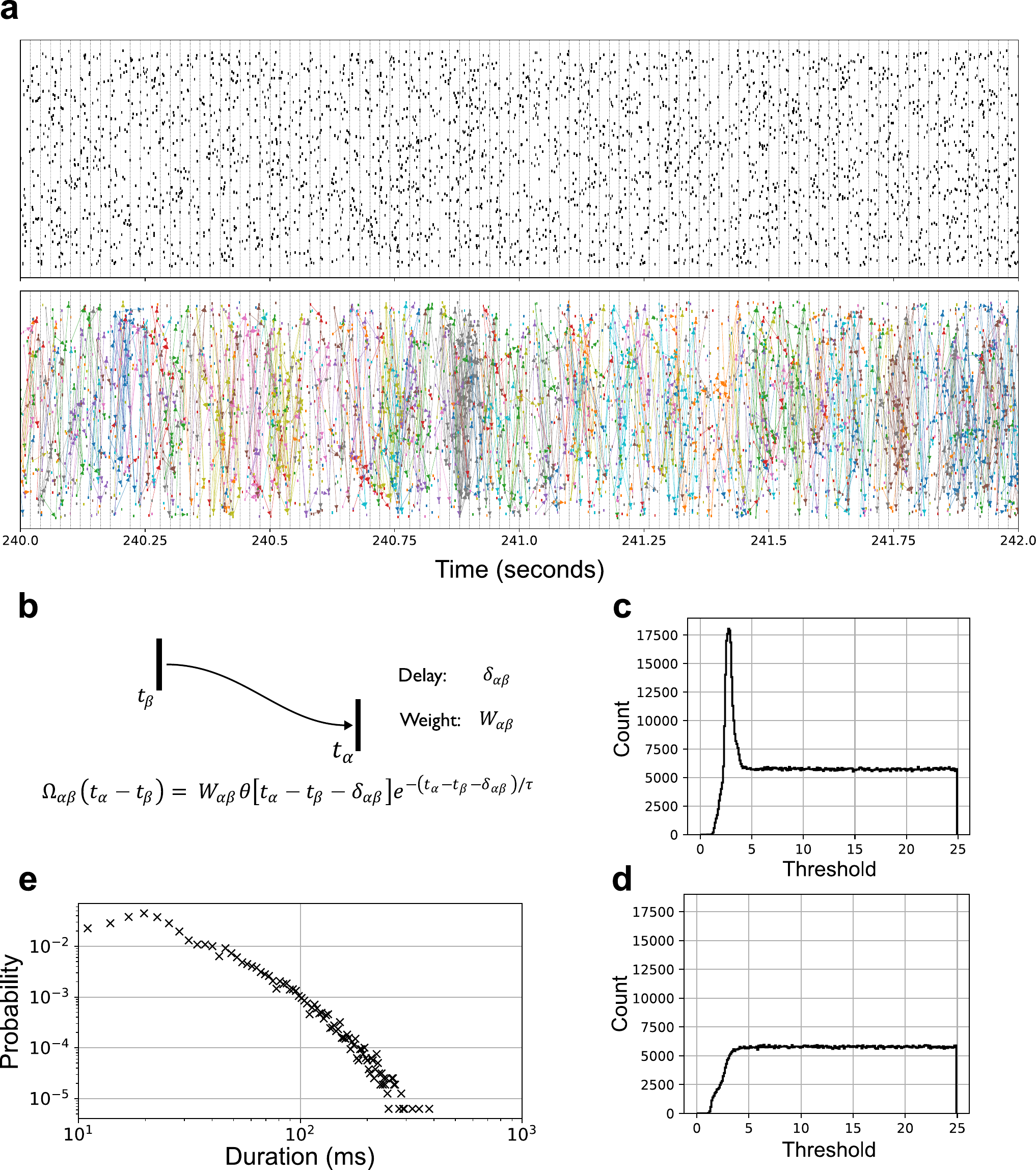}
\caption{Graphical neural activity threads. a) (top) Spike raster from 4000 excitatory cells in a geometrically-structured network freely evolving with spike-timing dependent plasticity for 10 minutes followed by 5 minutes of evolution with fixed synaptic weights. (bottom) The same spike train superimposed with directed edges representing strong causal relations between pairs of spikes. Edges and spikes are colored according to their membership in disjoint connected components. Vertical lines indicate 20ms time bins. b) Illustration of the causal relation computation. c) Distribution of $-\log \Omega$ for all pairs of spikes from a 5 minute recording. d) Same as c, but for shuffled spike trains. The peak is eliminated, indicating $\Omega$ is capturing significant synaptic interactions in the network. e) Duration distribution for GNATs from the simulation depicted in a.}
\label{fig:gnats}
\end{figure}

The causal action of presynaptic spikes on postsynaptic neurons is the fundamental interaction that defines spiking network dynamics. Thus, information processing in spiking networks must be supported on the causal relations between individual spikes. Synaptic interactions between neurons define a mathematical relation between spikes, and directed graphs usefully describe the structure of these relations. We seek to construct a graph out of spiking neural activity that captures these causal relations by defining a quantity that estimates the degree of causal relatedness between spikes. 

When an excitatory spike arrives at a postsynaptic neuron, its synaptic action brings the postsynaptic neuron closer to the spiking threshold by raising its membrane potential. The synaptic weight determines the magnitude of this increase, so the degree of causal relatedness depends directly on the weight. Postsynaptic potentials are not instantaneous, but decay in time. Likewise, the causal action of a presynaptic spike on a postsynaptic neuron lingers beyond the arrival of the presynaptic spike, and can interact with other presynaptic spikes to ultimately push the postsynaptic neuron beyond the spiking threshold. This means any quantitative estimate of the causal influence of presynaptic spikes should include this temporally decaying influence.

We define a quantity, $\Omega$, that captures these intuitions about the causal action of spikes in a single real value associated to a pair of spikes:

\begin{equation}
 \Omega(t_{\alpha}, t_{\beta}) = \frac{W_{\alpha \beta}}{||W||} \theta[t_\alpha - t_\beta - \delta_{\alpha \beta}] e^{\frac{-(t_\alpha - t_\beta - \delta_{\alpha \beta})}{\tau}}
\end{equation}

Here, $t_\alpha$ and $t_\beta$ are the times of the postsynaptic and presynaptic spikes, respectively. $W_{\alpha \beta}$ is the synaptic weight from neuron $\beta$ to neuron $\alpha$, and $||W||$ is the norm of all synaptic weights onto neuron $\alpha$. $\delta_{\alpha \beta}$ is the conduction from neuron $\beta$ to neuron $\alpha$. $\theta$ represents the Heaviside step function, ensuring that $\Omega$ is nonzero only for spike pairs that arise from synaptically connected neurons and are separated by at least one axonal conduction delay. The exponential term captures the decay of causal influence as the temporal separation of the spikes increases, with $\tau$ a free parameter that determines the rate of decrease. 

We define a directed acyclic graph we call the \emph{activity graph} from spike trains by computing $\Omega$ for each spike pair from excitatory neurons. Spike pairs with large $\Omega$ correspond to a significant causal relation between the spikes. We include a directed edge from a presynaptic spike to a postsynaptic spike in the activity graph if $\Omega$ exceeds a threshold, explained below. Because of the exponential decay, we can limit our computation to spike pairs that are separated in time by less than a few time constants, significantly improving the efficiency of our construction on large spike trains. 

Fig. \ref{fig:gnats}a, top, shows a spike train from 4000 excitatory neurons over two seconds in a simulated spiking network. Fig. \ref{fig:gnats}a, bottom, shows the activity graph constructed from the spikes above. This directed acyclic graph decomposes into disjoint weakly connected components, which we illustrate by coloring each disjoint component separately. A weakly connected component in a directed graph is a set of vertices that are maximally connected when the direction of the edges is forgotten. We refer to the disjoint weakly connected components of the activity graph as Graphical Neural Activity Threads (GNATs). 

We found that the negative logarithm of $\Omega$ displays a strong peak at large $\Omega$ values (Fig. \ref{fig:gnats}c, $\tau$ = 5ms), that disappears for shuffled spike trains (Fig. \ref{fig:gnats}d), indicating $\Omega$ successfully quantifies information transmission by the causal action of presynaptic spikes on postsynaptic neurons. This property also defines a natural threshold for defining the edges in our activity graph: we choose our threshold to correspond to the transition between the peak and flat parts of the distribution of $-\log \Omega$, here, approximately 5. If $\tau$ changes in the definition of $\Omega$, then the peak moves to a new position, but the shape of the distribution stays the same. Changing the threshold to match the new peak in the new distribution ensures that the edges included are independent of the specific values of $\tau$ and the threshold. Thus, our definition of the activity graph corresponds to the intrinsic activity of the network. 


The GNATs that emerge from spiking network activity show how individual spikes contribute to the global network dynamics. Interestingly, we found that spikes belonging to the same time bin often belong to causally-disjoint threads. If the causal action of any two spikes eventually converges, those spikes will belong to the same GNAT, by definition. Therefore, these disjoint spikes do not jointly contribute to downstream effects. Thus, such spikes are likely computationally independent (though, disjoint spikes may interact through inhibition). We also observed many isolated spikes, likely due to the random and patterned stimulation (see section \ref{sim_details}), that do not causally contribute to any GNAT. These spikes are likely computationally irrelevant, but this is not obvious without the perspective provided by the GNATs. 

\section{Finding Analogous Threads}

To ground a computational interpretation of the GNATs, we must define how to compare GNATs, and in particular, identify when similar GNATs reoccur. Indeed, identifying recurring neural sequences is a major research direction in contemporary neuroscience and is fundamental to current theories of memory (e.g. hippocampal replay \cite{carr_hippocampal_2011}) and spatiotemporally structured natural behaviors (e.g. birdsong \cite{hahnloser_ultra-sparse_2002}). Typically, recurring neural sequences are identified through comparing absolute spike times, which are defined by an external clock. Time warping studies have shown that neural activity that looks unstructured with respect to an external clock nevertheless contains significant structure that exists on flexible timescales across trials \cite{williams_discovering_2020}. Because an external clock has no intrinsic relevance to the neural circuit, it is unreasonable to assume that repeat sequences must occur with respect to absolute spike times. We construct an alternative definition of neural sequences using the intrinsic causal relations of the spiking activity captured by the GNATs.

\begin{figure}[h!]
\includegraphics[width=1\textwidth]{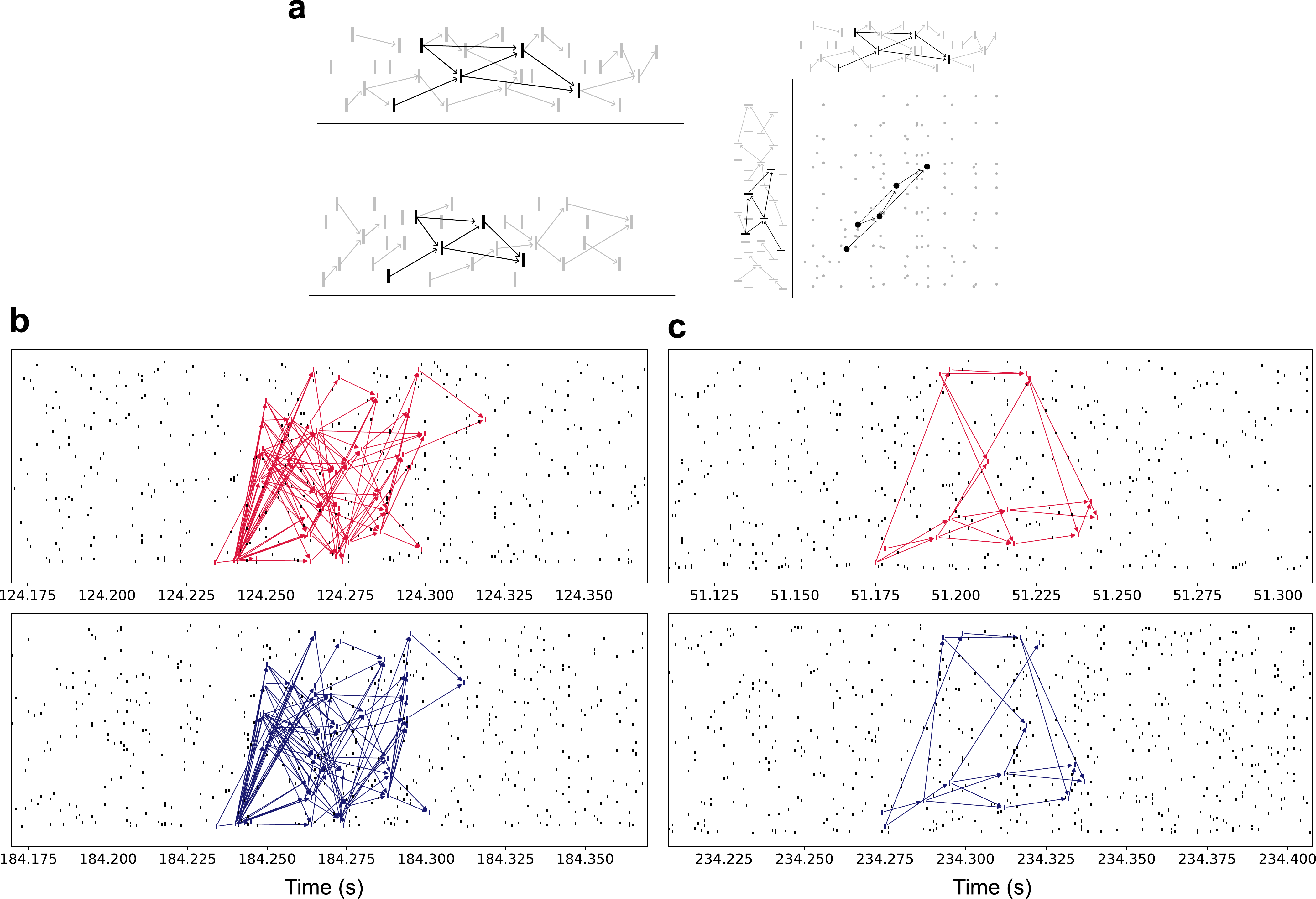}
\caption{Extracting analogous subthreads. a) Given two activity graphs, the second order activity graph is built out of the Cartesian product of spikes. Connected components in the second-order graph correspond to analogous subthreads. b) Example of an analogous subthread extracted from a 5-minute simulation of a 5000-neuron spiking network. Spikes with analogous causal relations reappear at different times in the simulation with slightly different timing relationships. c) Another example of an analogous subthread from the same simulation.}
\label{fig:iso_gnats}
\end{figure}

Because the relevant relation between spikes is causal rather than temporal, we define repeat neural sequences as repeated isomorphic subgraphs of the causal activity graph. Two isomorphic subgraphs identify spikes produced through the same synaptic interactions. For arbitrary graphs, the modular product \cite{barrow_subgraph_1976} is a graph product $G \times H$ defined for two graphs $G = (V_G, E_G)$ and $H = (V_H, E_H)$ as the graph with vertex set 
\begin{equation}
 V_{G \times H} = V_G \times V_H
\end{equation}
and edge set 
\begin{equation}
 E_{G \times H} = \left\{(u, v) \to (u', v') \, | \, [(u, u') \in E_G \land (v, v') \in E_H] \lor [(u, u') \notin E_G \land (v, v') \notin E_H] \right\}
\end{equation}
In other words, the vertex set of $V_{G \times H}$ is the cartesian product of $V_G$ and $V_H$, and the edge set is the exclusive NOR of $E_G$ and $E_H$. 
Importantly, cliques in the modular product correspond to isomorphic induced subgraphs between $G$ and $H$ \cite{barrow_subgraph_1976}. Thus, if we could construct the modular product of our neural activity graphs, cliques would identify isomorphic causal neural sequences. Enumerating cliques in graphs is hard in general, and the modular product is never sparse, so this approach for finding isomorphic neural activity subthreads is computationally intractable. Instead, we define a weaker modular product that is efficiently computable but still extracts causally-similar neural sequences.

\begin{figure}[h!]
\includegraphics[width=1\textwidth]{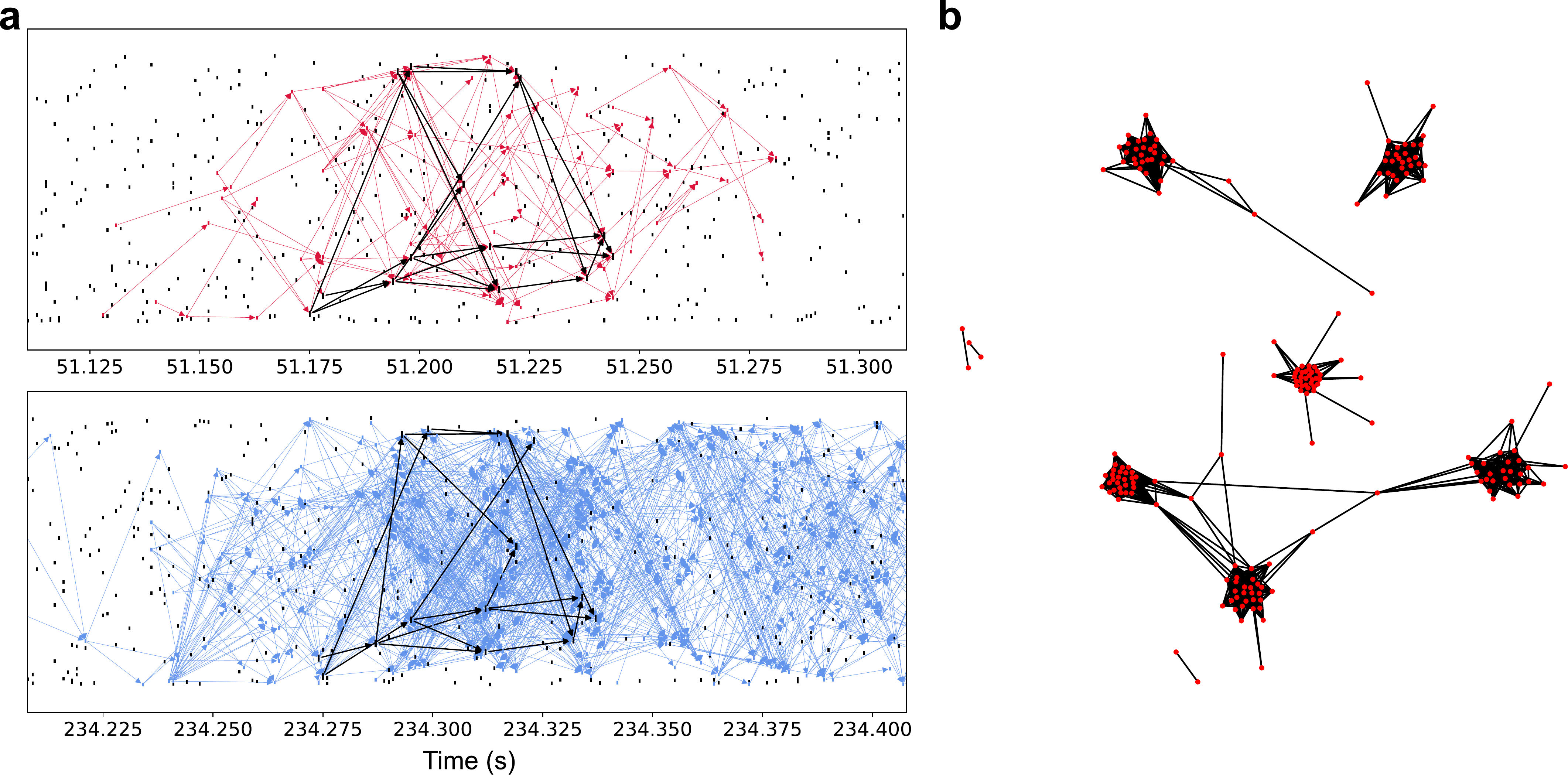}
\caption{Relationships between GNATs induced by analogous subthreads. a) The same analogous subthread as Fig \ref{fig:iso_gnats}c, shown embedded in their maximal GNATs. Each analogous subthread links two disjoint threads. b) Multigraph showing the relation between threads induced by the largest 2000 analogous subthreads. Each vertex corresponds to a maximal GNAT (i.e. as depicted by distinct colors in Fig. \ref{fig:gnats}a), and edges correspond to an analogous subthread. This graph of GNAT relations shows strong clustering, naturally partitioning the GNATs into classes. The graph is embedded with a spring layout with edge weights given by the number of spikes shared between subthreads. Parallel edges in the multigraph are not shown for clarity. }
\label{fig:gnat_relations}
\end{figure}

Given two activity graphs $A_i$ and $A_j$, we define a second order activity graph as follows. Let $A_{i, n}$ correspond to the set of spikes from neuron $n$ in activity graph $i$. The vertex set of the second order activity graph associated to activity graphs $A_i$ and $A_j$ is given by 
\begin{equation}
 V_{A_i \times A_j} = \bigcup_{n} A_{i, n} \times A_{j, n}
\end{equation}

and edge set given by 
\begin{equation}
 E_{A_i \times A_j} = \left\{ (u_a, u_b) \to (v_a, v_b) \, | \, (u_a, v_a) \in E_{A_i} \land (u_b, v_b) \in E_{A_j}\right\}
\end{equation}

Restated, the second order vertices are pairs of spikes from individual cells, and second order edges correspond to pairs of edges that both appear in the first order activity graph. Thus, a second order edge corresponds to the reappearance of a causal edge in the first order activity graph. With this definition, repeat causal neural sequences correspond to weakly connected components in the second order activity graph (Fig. \ref{fig:iso_gnats}a). These weakly connected components do not always correspond to exactly isomorphic subgraphs, so to avoid confusion, we refer to these second-order connected components as \textit{analogous subthreads}.

Because of the locality in space and time of causal edges, we can efficiently construct the second order activity graph by partitioning the cartesian product of each neuron's spikes using a quadtree data structure, and using the efficient spatial indexing in quadtrees to quickly find putatively connected pairs of spikes in the second order activity graph. 

We applied our extraction algorithm to the spike trains generated by our example network (Sec. \ref{sim_details}), and found significant numbers of analogous subthread pairs (n=857919 in a 5-minute simulation). Most of these pairs consisted of less than 15 spikes, but we extracted $\sim 2000$ analogous thread pairs containing 15 or more spikes. Depending on the simulation, analogous thread pairs contained a maximum of 90 to several hundred spikes. Figs. \ref{fig:iso_gnats}b and \ref{fig:iso_gnats}c show example analogous thread pairs found by our algorithm. The analogous threads in this example contain similar causal relations between spikes, even though the absolute timing relationships between these patterns are not preserved. 
\section{Relations between GNATs}

Analogous subthreads are embedded in larger, maximally connected GNATs by definition. Fig. \ref{fig:gnat_relations}a shows the pair of analogous subthreads from Fig. \ref{fig:iso_gnats}c embedded in their larger contexts. In this way, individual GNATs can be thought of as composed of subthreads that can reappear in distinct GNATs. To understand the compositional structure of GNATs induced by analogous subthreads, we constructed a multigraph with disjoint, maximally-connected GNATs (Fig. \ref{fig:gnats}a) as vertices and edges between pairs of GNATs corresponding to the analogous subthreads (Fig. \ref{fig:iso_gnats}b,c) that appear in both threads of the pair. Fig. \ref{fig:gnat_relations}b shows an example of this GNAT composition graph including the 2000 largest analogous subthreads found in a 5-minute simulation. We found that individual GNATs strongly cluster into distinct classes based on their shared subthreads. 

\begin{figure}[h!]
\includegraphics[width=1\textwidth]{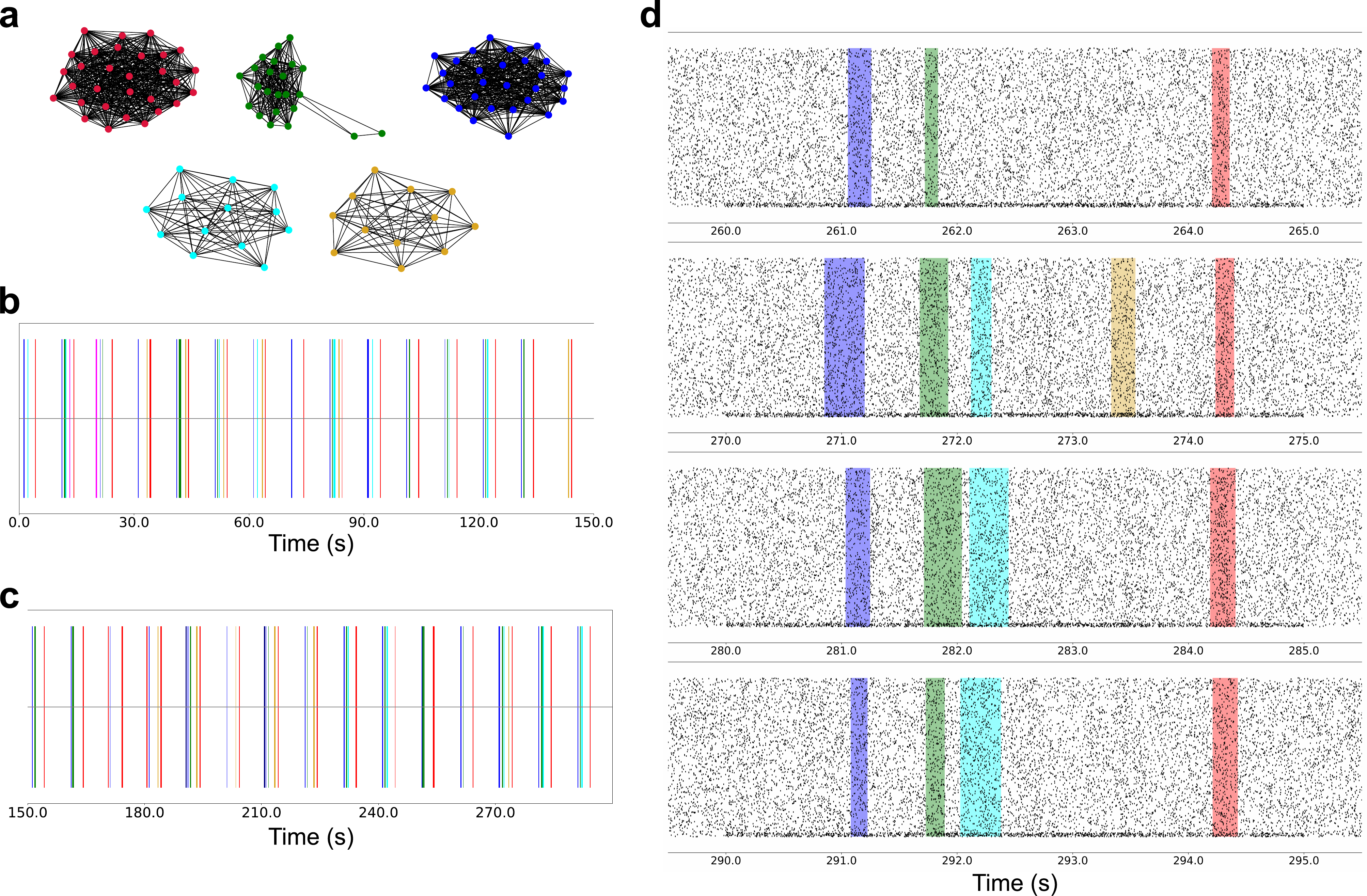}
\caption{GNATs reveal compositionality of spiking responses. a) Individual GNAT classes extracted from the graph in Fig. \ref{fig:gnat_relations}b. b,c) Intervals during the 5-minute simulation corresponding to GNAT classes from a. The sequence of GNAT classes shows regularity and flexibility simultaneously. d) Intervals corresponding to GNAT classes overlaid on spiking responses to a fixed input pattern. The spiking response is highly variable trial to trial, but GNATs embedded within each response are composed of analogous causal sequences not apparent from an inspection of solely the temporal relations between spikes.}
\label{fig:gnat_trials}
\end{figure}

We extracted these classes of GNATs and plotted the temporal extent of each member of each class on top of the original spike train. Fig. \ref{fig:gnat_trials}a shows the individual GNAT classes and their assigned colors. Fig. \ref{fig:gnat_trials} b and c show the temporal extent of each GNAT colored according its class membership over a five minute simulation. Fig. \ref{fig:gnat_trials}d shows the temporal extent of GNATs belonging to each class over four trials of repeated stimulation according to a fixed input spike pattern. Like-colored intervals indicate the temporal extent of GNATs embedded in the spiking activity belonging to the same class, meaning they share analogous subthreads. The network's response to the spike pattern on each trial shows regularity and flexibility in the appearance of GNATs of each class. While this response is variable, it is composed of activity threads built from analogous causal sequences. This result highlights the benefits of the GNAT approach in capturing the challenging balance between between flexibility and rigidity in spiking networks.

\begin{figure}[h!]
\includegraphics[width=1\textwidth]{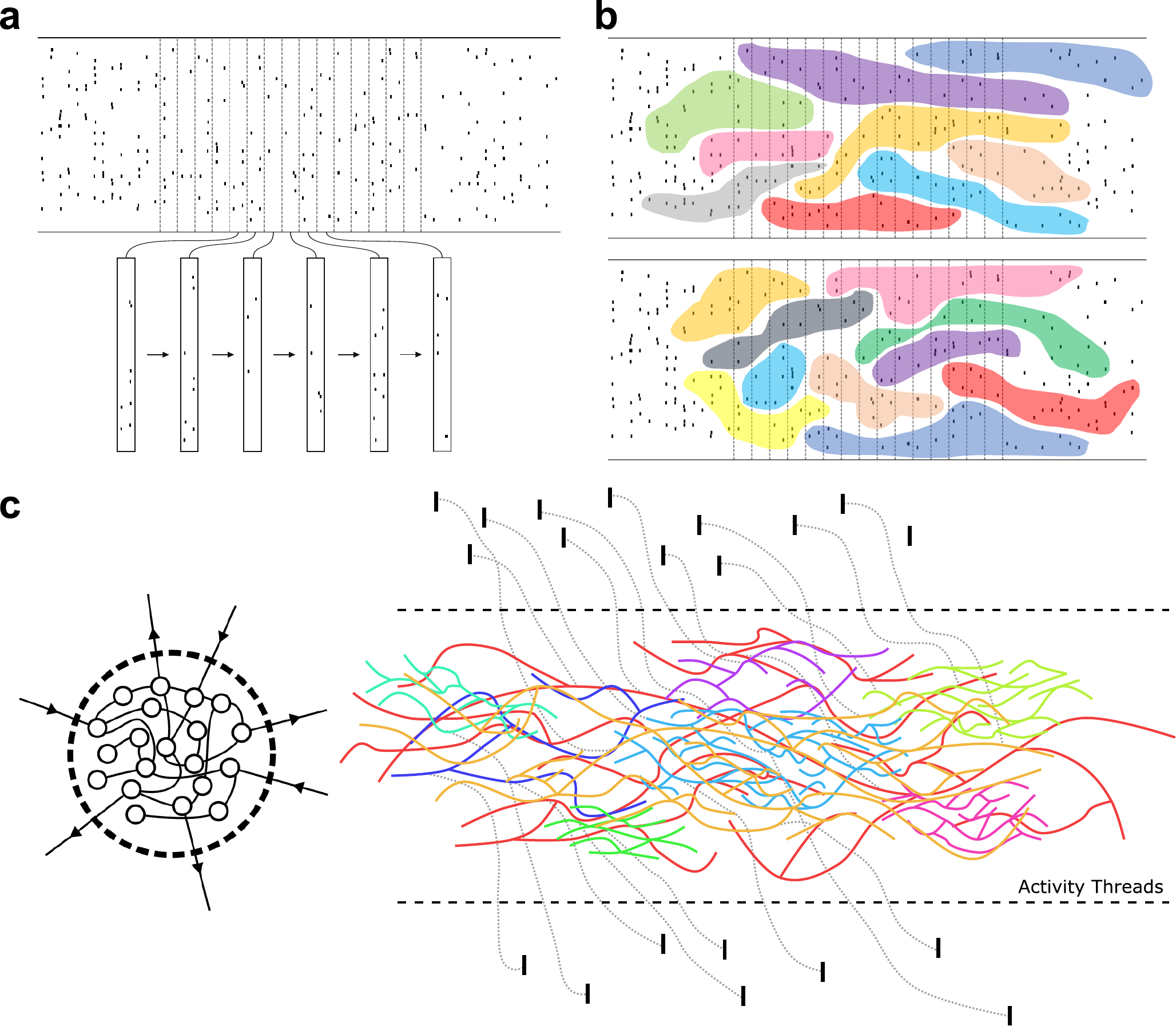}
\caption{Computational abstractions for spiking neural networks from GNATs. a) Typical spiking analyses require partitoning time into bins. Simultaneous spikes in each bin form a distributed representation of computational variables in an effectively serial computation. b) Similar spike time patterns could emerge from distinct underlying causal processes. These distinctions are not captured through the state vectors but are captured by GNATs incorporating spike times and network connectivity in a single structure. c) Efferent and afferent spikes in spiking networks interact with ongoing activity threads to determine the overall computation. GNATs allow for tracing the causal history of each spike from output back to input, providing a detailed account of spiking computations.}
\label{fig:gnat_abstractions}
\end{figure}

\section{Conclusions}
We have shown that spiking neural activity naturally decomposes into disjoint, discrete threads. These threads are composed of causal sequences that can be reused and flexibly recombined, and we have provided an algorithm that extracts analogous subthreads from larger spike trains. As such, GNATs are promising candidates for elementary computations in spiking networks. They exhibit properties such as spatial and temporal parallelism, key intuitions of spiking neural computation inherited from neuroscience that have so far resisted a rigorous definition. 

Our definition of GNATs avoids many of the artificial assumptions fundamental to traditional abstractions of spiking neural computation. In many approaches, time is partitioned into bins according to the clock of an external observer. Simultaneous neural activity in each bin defines a population state vector that represents the computational variables, and the spiking computation is given by the sequence of state vectors (Fig. \ref{fig:gnat_abstractions}a). In this scheme, the state-to-state transition is computed in parallel, but the effective computation is still a serial sequence of state vectors. GNATs are fully asynchronous -- they require no time bins, and more than one GNAT can exist over any time interval. Furthermore, state vector schemes assume that spikes belonging to the same bin belong to the same computation. Our GNAT analysis reveals that spikes belonging to the same time bin may belong to causally-disjoint sequences of activity (Fig \ref{fig:gnats}a, bottom). If two spikes belonged to the same computation, then their downstream causal effects should eventually converge, meaning that they would belong to the same GNAT under our definition. 

Our GNAT analysis combines the temporal structure of spike trains with the connectivity structure of the neural network in a single object -- the activity graph. Most spiking analyses work without connectivity information, but the connectivity is what gives meaning to spiking activity. Fig. \ref{fig:gnat_abstractions}b shows two temporally-similar spike trains. The two similar spike trains could emerge from distinct networks. In each case, the underlying synaptic interactions determining the temporal dynamics are distinct, illustrated by the distinct partitions of the top and bottom spike trains. These distinctions would be transparent to a population state vector analysis, but are intrinsic to the definition of the GNATs. 

The picture of spiking computation that emerges from the GNAT point of view is depicted in Fig. \ref{fig:gnat_abstractions}c. A recurrent spiking network sends and receives spikes to and from its external environment. These afferent/efferent spikes interact with ongoing activity threads to specify the computation. Our GNAT analysis allows for tracing the causal history of every spike, illustrating in detail how spiking networks transform their afference to efference. 

Despite the appealing properties of GNATs we observed, we have not connected GNATs to a specific neural computation. To do so will require a thorough understanding of GNATs and their interactions. Due to the combinatorially large number of possibile patterns of activity produced by spiking networks, this is a challenging problem. However, our initial results indicate that there is significant structure to the pattern of GNATs that emerge from a spiking network (Fig. \ref{fig:gnat_relations}b), so there is likely to be rich theory underlying GNAT dynamics. Furthermore, the connectivity of a network places strong constraints on possible threads, so a satisfactory theory of GNATs will further our understanding of the relation between spiking network structure and dynamics, an important research direction in neural computing. $\Omega$ is strongly related to STDP facilitation kernels, which suggests interpreting edges in the causal activity graph as potential synaptic plasticity events. This directly links network activity and network connectivity: the connectivity determines potential GNATs, and the actualized GNATs determine the evolution of the connectivity. Future work will explore this connection in a learning context.

\section{Simulation details}
\label{sim_details}
We simulated spiking networks using the STACS spiking network simulator \cite{stacs_2015}. Networks had 4000 excitatory neurons (Izhikevich RS) and 1000 inhibitory neurons (Izhikevich FS) \cite{izhikevich_simple_2003}. Neurons were randomly distributed on a $100 \mu m \times 100 \mu m$ periodic rectangle (a torus), and connectivity was randomly initialized according to a probability that varied with the physical distance between neurons. 
\begin{equation}
 p(r)= P_{\textsc{max}} \left( 1 - \frac{1}{1+\exp (-\sigma (r-\mu))} \right)
\end{equation}
E to E,I connections used $P_{\textsc{max}} = 0.4$, $\mu = 10$, $\sigma = 1$ and I to E connections used $P_{\textsc{max}} = 0.5$, $\mu = 10$, $\sigma = 1$. Excitatory conduction delays were chosen uniformly randomly in the range $[1 ms, 20 ms]$ and inhibtory delays were fixed at $1 ms$. 
The networks evolved for 10 simulated minutes and weights were allowed to change according to an STDP rule \cite{izhikevich_polychronization_2006}. Then, the weights were fixed and the network evolved for 5 minutes under the same stimulation conditions. GNATs were computed from this 5 minute simulation.
Networks were driven with Poisson spikes applied to each neuron independently at a rate of 0.4 Hz. On top of this random stimulation, a randomly chosen but fixed spike pattern was applied to 100 randomly chosen neurons throughout the plastic and fixed phases of the simulation. The spike pattern consisted of random spikes at a rate of 2 Hz distributed over a 5 second interval, followed by a 5 second period of silence. Thus, networks experienced 60 repetitions of the pattern during the plastic phase and 30 repetitions during the fixed phase. 

\begin{ack}
This material is based upon work supported by the U.S. Department of Energy, Office of Science, Office of Advanced Scientific Computing Research (ASCR) as part of the Collaborative Research in Computational Neuroscience Program.

This article has been authored by an employee of National Technology \& Engineering Solutions of Sandia, LLC under Contract No. DE-NA0003525 with the U.S. Department of Energy (DOE). The employee owns all right, title and interest in and to the article and is solely responsible for its contents. The United States Government retains and the publisher, by accepting the article for publication, acknowledges that the United States Government retains a non-exclusive, paid-up, irrevocable, world-wide license to publish or reproduce the published form of this article or allow others to do so, for United States Government purposes. The DOE will provide public access to these results of federally sponsored research in accordance with the DOE Public Access Plan https://www.energy.gov/downloads/doe-public-access-plan.
This paper describes objective technical results and analysis. Any subjective views or opinions that might be expressed in the paper do not necessarily represent the views of the U.S. Department of Energy or the United States Government. SAND2023-05685O
\end{ack}


\bibliographystyle{plain}
\bibliography{gnats}

\end{document}